%% file: main.tex
\documentclass{article}

\PassOptionsToPackage{numbers,compress}{natbib}

\usepackage[preprint]{neurips_2026}

\usepackage[utf8]{inputenc}
\usepackage[T1]{fontenc}
\usepackage{hyperref}
\usepackage{url}
\usepackage{booktabs}
\usepackage{amsfonts}
\usepackage{nicefrac}
\usepackage{microtype}
\usepackage{xcolor}
\usepackage{graphicx}
\usepackage{amsmath}
\usepackage{amssymb}
\usepackage{array}
\usepackage{tabularx}
\usepackage{multirow}
\usepackage{makecell}
\usepackage{bm}
\usepackage{enumitem}

\newcommand{\methodname}{LoD-Loc v4}

\title{\methodname: Towards Generalized Aerial Visual Localization via Panoptic LoD 3D Map Registration}

\author{%
  \textbf{Shuaibang Peng} \quad
  \textbf{Chengming Zhang} \quad
  \textbf{Juelin Zhu} \quad
  \textbf{Xia Li} \quad
  \textbf{Xiaoya Cheng} \\[2pt]
  \textbf{Zeyu Cui} \quad
  \textbf{Shen Yan}\thanks{Correspondence to
    \texttt{yanshen12@nudt.edu.cn}.} \quad
  \textbf{Maojun Zhang} \quad
  \textbf{Yu Liu} \\[2pt]
  \normalfont National University of Defense Technology
}

\begin{document}

\maketitle

\input{section/abstract.tex}
\input{section/introduction.tex}
\input{section/related_work.tex}
\input{section/method.tex}
\input{section/airpod_dataset.tex}
\input{section/experiments.tex}
\input{section/conclusion.tex}
\section*{Acknowledgments}
This work was supported by the National Natural Science
Foundation of China under Grant No.~62406331.


\bibliographystyle{abbrvnat}
\bibliography{main}

\appendix
\input{section/appendix.tex}


\end{document}

%% file: section/abstract.tex
\begin{abstract}
Aerial visual localization over Level-of-Detail (LoD) 3D Maps has recently gained significant traction due to its global scalability, storage efficiency, and privacy-preserving nature. However, existing methods rely almost exclusively on building elements, which leads to catastrophic localization failures in rural or undeveloped areas where buildings are sparse or non-existent. 
To address this, we construct a unified Panoptic LoD 3D Map by fusing OpenStreetMap (OSM), Digital Elevation Models (DEM), and LoD assets, reformulating the problem as a panoptic registration task. By parsing the query image into discrete building `thing' instances and continuous `stuff' regions comprising road, vegetation, and water, we establish a holistic alignment with this map to robustly optimize the camera pose. 
To enable reliable panoptic perception, we develop a new automated pipeline that constructs \textbf{AirPan} - the largest aerial panoptic segmentation dataset to date, comprising 85k images with pixel-accurate annotations. This empowers our panoptic segmentation model to exhibit remarkable zero-shot generalization capabilities. 
Extensive experiments demonstrate that LoD-Loc v4 consistently outperforms state-of-the-art (SOTA) baselines by a significant margin. Most notably, it successfully maintains reliable localization in building-less scenarios, completely overcoming the catastrophic failures of prior building-centric paradigms.
\end{abstract}

%% file: section/introduction.tex
\section{Introduction}
\label{sec:intro}

\begin{figure*}[htbp]
    \centering
    \includegraphics[width=\linewidth]{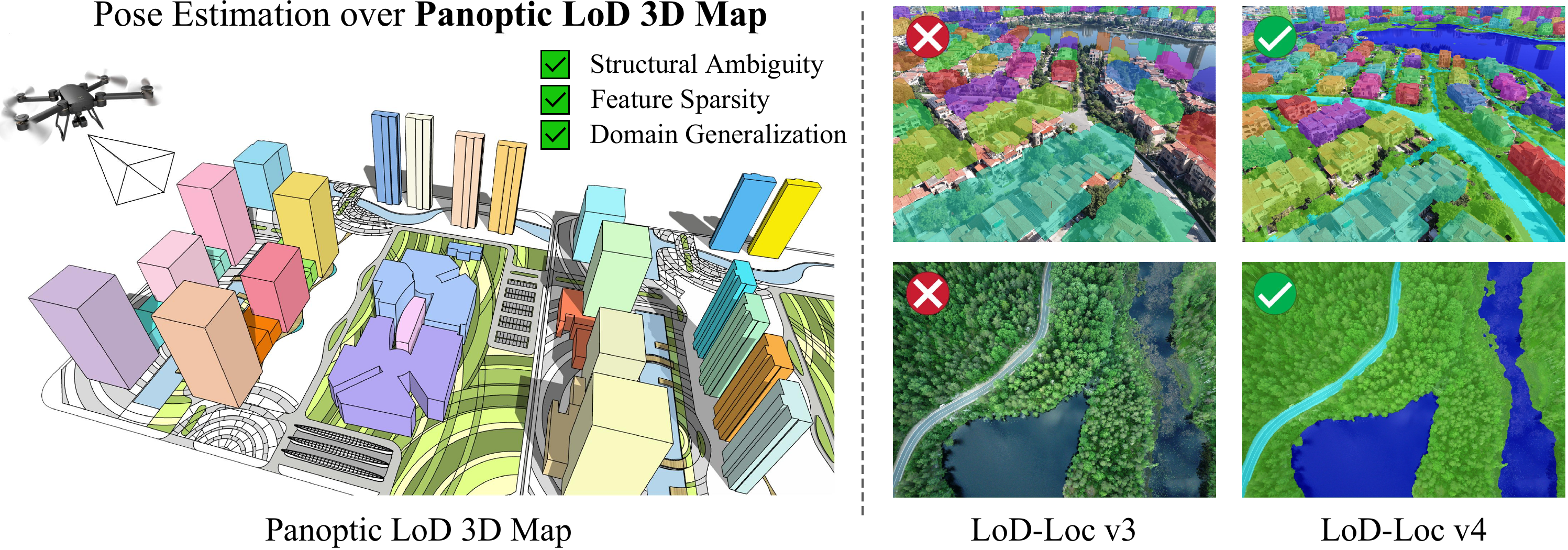}
    \caption{In this paper, we introduce LoD-Loc v4 to extend aerial localization beyond building-centric scenes based on the Panoptic LoD 3D Map. In urban scenes, our method outperforms SOTA baselines. In building-less scenes, \methodname\ maintains reliable results by utilizing road, vegetation, and water layouts as panoptic cues, whereas previous methods fail entirely.}
  \label{fig:teaser_intro}
\end{figure*}

Aerial visual localization, the task of estimating the 6-DoF pose of an unmanned aerial vehicle (UAV) against a geo-referenced map, is a foundational technology for critical applications, including GNSS-denied navigation, UAV-based VR/AR, and digital twin systems.

Conventional methods localize UAVs using 2D geographic priors, primarily DEM and Digital Orthophoto Maps (DOM)~\cite{kinnari2022season, zheng2020university, chen2021real, shi2020looking, shi2022beyond}. These methods typically solve the camera pose by performing feature matching between the UAV query image and the reference map, followed by the calculation of geometric transformations such as similarity or homography. Despite its prevalence, this 2D reference-based approach encounters a fundamental limitation when the UAV operates at lower altitudes or with oblique viewing angles, as standard 2D maps completely lack view-dependent building height information, leading to localization failure.

To bridge this gap, recent methods investigate localization against photorealistic 3D maps, such as mesh models derived from oblique photogrammetry. These approaches align sparse~\cite{yan2023render, panek2022meshloc} or dense query features~\cite{cheng2026pilotneuralpixelto3dregistration} with corresponding features synthesized from the reference mesh. While effective across diverse view angles and altitudes, the acquisition cost, storage overhead, and maintenance burden of photorealistic 3D maps are prohibitive for large-scale deployment. Furthermore, this map raises privacy concerns for individuals and the government.

We therefore pose a question: \textit{Can we find a 3D map that is global in scale, lightweight in storage, and privacy-compliant, while capable of supporting UAV localization across extreme altitudes and oblique viewing angles?} Panoptic LoD 3D Maps, the "white-box" models in commercial navigation apps (i.e., Google Maps, Apple Maps, and AMap), provide a highly practical and promising approach. These maps decompose complex scenes into structured semantic layers (e.g., road, water, vegetation) and discrete instance geometries (e.g., buildings). Although preliminary studies~\cite{zhu2024lod, zhu2025lod, peng2026lodlocv3} have attempted to utilize buildings from such maps for visual localization, their effectiveness is fundamentally bottlenecked by the visibility of building instances, leading to complete localization failure in scenes where building structures are sparse or absent.

To address this, we propose a \textbf{Panoptic Registration Paradigm} that shifts the focus from purely geometric building matching to a holistic alignment of all panoptic elements. Our strategy integrates discrete instance-level `thing' (e.g., buildings) with continuous `stuff' (e.g., road, vegetation, water), yielding a dense and robust geometric signal for pose estimation. To enable the zero-shot capability of panoptic segmentation for query, we introduce \textbf{AirPan}, a massive synthetic dataset for aerial panoptic segmentation. Specifically, we develop a highly automated pipeline that fuses OSM, DEM, and LoD assets to generate 85k synthetic images with pixel-level labels, eliminating manual intervention and far exceeding existing benchmarks. As illustrated in Figure~\ref{fig:teaser_intro}, \methodname\ achieves successful localization in building-less regions where prior studies catastrophically fail. Moreover, our approach yields substantial accuracy gains compared to baselines even in urban areas.

Our main contributions are summarized as follows:
\begin{itemize}[leftmargin=1.2em]
  \item We construct a Panoptic LoD 3D Map and propose a new aerial localization paradigm that reformulates 4-DoF pose estimation as panoptic registration over this map.

  \item We develop a scalable automated simulation pipeline and introduce \textbf{AirPan}, to the best of our knowledge, the largest and most diverse aerial panoptic segmentation dataset.
  \item We demonstrate through extensive experiments that our method consistently outperforms baselines by a large margin, especially in building-less scenes.
\end{itemize}

%% file: section/related_work.tex
\section{Related Work}
\label{sec:related}

\noindent \textbf{Aerial visual localization.} Aerial visual localization estimates a UAV's 6-DoF pose by aligning query images with geo-referenced map priors~\cite{couturier2021review, couturier2024review, piasco2018survey}. Conventional 2D pipelines using DEM and DOM~\cite{kinnari2022season, zheng2020university, chen2021real, shi2020looking, shi2022beyond} estimate planar transformations but fail at low altitudes or oblique angles due to missing 3D height cues. To address this, high-precision methods utilize photorealistic 3D maps like Structure-from-Motion (SfM) point clouds~\cite{schonberger2016structure, wu2024uavd4l} and meshes~\cite{panek2022meshloc, yan2023render}. They align sparse~\cite{yan2023render, panek2022meshloc} or dense~\cite{cheng2026pilotneuralpixelto3dregistration} features via local matching~\cite{lowe2004distinctive, sarlin2020superglue, sun2021loftr} and solve poses via PnP-RANSAC~\cite{haralick1994review, kneip2011novel, barath2019magsac}. Though effective, these maps are computationally heavy, costly to scale, and pose privacy risks. Alternatively, publicly available maps such as OSM support localization without photorealistic 3D reconstruction~\cite{sarlin2023orienternet, sarlin2024snap, Pietrantoni_2023_CVPR}. OrienterNet~\cite{sarlin2023orienternet} matches a neural bird's-eye view with rasterized OSM maps, while VecMapLocNet~\cite{wang2025vecmaplocnet} estimates planar location and yaw through cross-modal matching between UAV imagery and OSM vector maps. LoD 3D Maps~\cite{groger2012ogc} additionally provide explicit 3D geometry, which existing LoD-based localization frameworks exploit primarily through building structures~\cite{zhu2024lod, zhu2025lod, peng2026lodlocv3}.

\noindent \textbf{Localization with geographic 3D maps.} The LoD-Loc series forms our direct foundation. LoD-Loc~\cite{zhu2024lod} proved the feasibility of aerial localization over 3D city models via neural wireframe alignment with LoD2/3 geometries. LoD-Loc v2~\cite{zhu2025lod} enhanced scalability using LoD1 maps and semantic building silhouettes, while LoD-Loc v3~\cite{peng2026lodlocv3} resolved dense urban ambiguities via instance-level matching. However, their exclusive reliance on building geometries restricts their operational scope. Beyond urban UAV localization, GeoPose3K~\cite{brejcha2017geopose3k} provides a mountain-landscape dataset for outdoor camera pose estimation. We transcend this building-centric bottleneck by introducing a unified \textbf{Panoptic LoD 3D Map}. This formulation preserves discriminative building structures as `thing' instances while explicitly incorporating continuous land-cover regions as `stuff' semantics. By jointly aligning these elements with the query image, our approach enables localization in the evaluated building-sparse and building-less environments.

\noindent \textbf{Aerial panoptic segmentation.} Panoptic segmentation unifies instance and semantic scene understanding~\cite{kirillov2019panoptic}, recently driven by transformer architectures like Mask2Former~\cite{cheng2022masked} and large vision models~\cite{kirillov2023segment, zou2024segment, zhao2023fast}. Deploying these to aerial imagery suffers from severe domain shifts. To avoid prohibitive retraining, recent works adopt parameter-efficient fine-tuning (PEFT) strategies like adapters and Low-Rank Adaptation (LoRA)~\cite{jia2022vpt, hu2022lora, chen2023vitadapter, chen2024rsprompter, yi2024learning, wei2024stronger, bi2024learning}. While we adopt this efficient paradigm using a low-rank adapted backbone and Mask2Former decoder, our objective diverges from generic aerial panoptic segmentation. Standard tasks exhaustively parse scenes, including transient classes (e.g., vehicles, pedestrians) that act as severe geometric noise during map matching~\cite{schonberger2018semantic, bescos2018dynaslam}. We circumvent this by distilling the scene into a localization-oriented Panoptic LoD 3D Map. We intentionally discard transient noise, strictly isolating stable building structures as `thing' instances and static environmental context as `stuff' regions to provide unambiguous geometric constraints.

\noindent \textbf{Aerial segmentation datasets.}
Current aerial semantic segmentation datasets~\cite{nigam2018ensemble, lyu2020uavid, rahnemoonfar2021floodnet, azimi2019skyscapes} provide essential semantic foundations but have two limitations: limited viewpoint and geographic diversity, and difficulty in linking images and camera poses to corresponding geo-referenced 3D maps, limiting their use for map-based aerial localization. To reduce annotation costs and introduce geometric priors, recent efforts utilize synthetic environments (e.g., SynDrone~\cite{rizzoli2023syndrone}, VALID~\cite{valid}, ClaraVid~\cite{claravid}) or map-assisted supervision~\cite{kaiser2017learning}. However, existing synthetic datasets often rely on fictional environments lacking real-world structural diversity, and map-assisted supervision has been explored using orthorectified aerial imagery~\cite{kaiser2017learning}. To address the above limitations, our \textbf{AirPan} dataset covers diverse geographic regions and viewing angles and associates images and camera poses with corresponding Panoptic LoD 3D Maps built from DEM, OSM, and LoD models. These maps are used for both panoptic annotation and localization, providing a common geographic reference for the two tasks.

%% file: section/method.tex
\section{Method}
\label{sec:method}

\begin{figure*}[htbp]
    \centering
    \includegraphics[width=\linewidth]{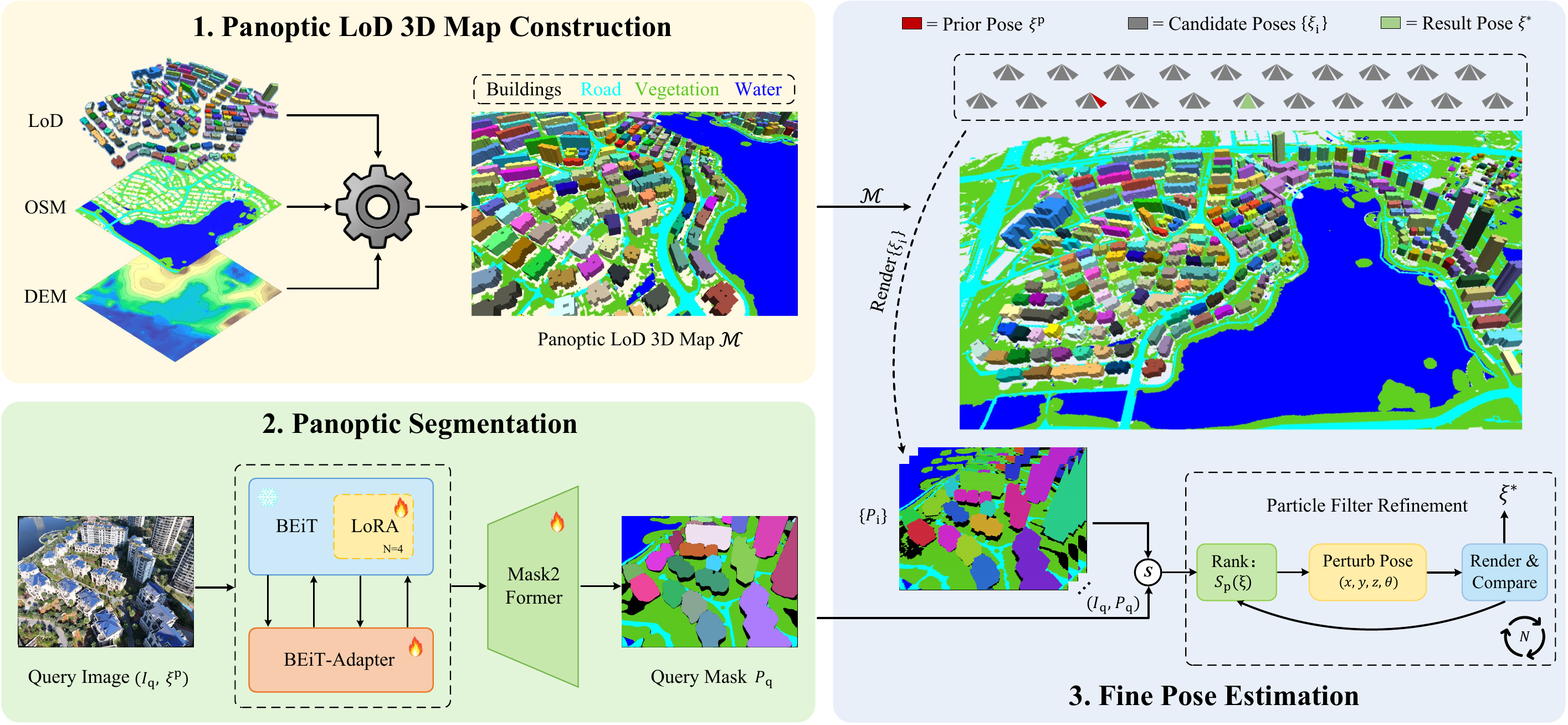}
    \caption{\textbf{Overview of \methodname.} 1. Constructing a Panoptic LoD 3D Map $\mathcal{M}$ from DEM, OSM, and LoD models (Section~\ref{sec:pod-map}). 2. Extracting building instances $\mathcal{B}_q$ and continuous regions $\mathcal{R}_q$ from query $I_q$ via a segmentation module (Section~\ref{sec:pod-seg}). 3. Refining the prior pose $\boldsymbol{\xi}^{p}$ to $\boldsymbol{\xi}^{*}$ via panoptic alignment between predicted masks and rendered hypotheses (Section~\ref{sec:localization}).}
  \label{fig:pipeline}
\end{figure*}

\noindent \textbf{Problem Formulation.}
Given an aerial query image $I_q$, a geographically aligned Panoptic LoD 3D Map $\mathcal{M}$, and a prior pose $\boldsymbol{\xi}^{p}$, our goal is to recover the absolute camera pose. Here, $\boldsymbol{\xi}^{p}$ denotes an externally supplied initial estimate of the camera pose for the current query. In our evaluation, it is obtained from onboard GPS/IMU measurements or synthesized by perturbing the ground-truth pose, depending on the benchmark (Section~\ref{sec:exp_setup}). Each query is localized independently, and its prior is not the localization output of the previous frame. Following established aerial localization paradigms~\cite{yan2023long, sarlin2023orienternet, sarlin2024snap, lynen2020large, Fragoso_2020_CVPR}, we leverage onboard inertial sensors that reliably provide the gravity direction to determine pitch and roll. This reduces the pose estimation problem to a 4-DoF $\boldsymbol{\xi} = (x, y, z, \theta)$, comprising 3D translation and 1D yaw.

\noindent\textbf{Overview.}
The \methodname\ pipeline contains three stages as shown in Figure~\ref{fig:pipeline}. First, we construct a Panoptic LoD 3D Map from publicly available geographic data in Section~\ref{sec:pod-map}. Second, we build a panoptic segmentation model that converts the query image into a panoptic mask compatible with the map representation in Section~\ref{sec:pod-seg}. Third, we build a particle-filter-based localization framework that refines the prior pose by evaluating the alignment between the predicted panoptic silhouette of the query image and rendered Panoptic LoD 3D Map hypotheses in Section~\ref{sec:localization}.

\subsection{Panoptic LoD 3D Map Construction}
\label{sec:pod-map}

Panoptic LoD 3D Map jointly encodes buildings as `thing' instances alongside road, vegetation, and water as `stuff' regions. To construct this map at a global scale, we design a fully automated pipeline that fuses three public data modalities: OSM~\cite{OSM}, DEM, and LoD models~\cite{peng2026lodlocv3}. 

The pipeline rasterizes OSM land-use polygons and aligns them with DEM tiles to generate a semantically labeled 2.5D terrain surface. We then inject topologically instantiated LoD building models onto this terrain. To support visual rendering, each building instance is assigned a unique random color, while `stuff' categories utilize fixed semantic colors. Formally, the resulting map is defined as $\mathcal{M} = \{\mathcal{B},\, \mathcal{R}\}$, where $\mathcal{B} = \{B_i\}_{i=1}^{N}$ denotes the set of $N$ discrete building instances (`thing' elements), and $\mathcal{R} = \{R^c\}_{c \in \mathcal{C}}$ denotes the continuous `stuff' regions spanning the semantic space $\mathcal{C} = \{\text{road}, \text{vegetation}, \text{water}\}$. This unified scene is ultimately stored in the OpenSceneGraph Binary (OSGB) format, with comprehensive generation details provided in Appendix~\ref{app:dataset}.

\subsection{Panoptic Segmentation Architecture}
\label{sec:pod-seg}

The panoptic segmentation module converts the aerial query image $I_q$ into a panoptic representation compatible with the rendered Panoptic LoD 3D Map. To jointly handle discrete instance-level discrimination and continuous semantic region coverage, we build upon the Mask2Former~\cite{cheng2022masked} framework combined with a BEiT-Adapter~\cite{chen2023vitadapter} backbone.

To overcome the substantial domain gap between natural pretraining imagery and our aerial domain without severe overfitting, we freeze the core BEiT encoder and introduce a PEFT strategy. Specifically, we inject LoRA modules~\cite{hu2022lora} into the attention layers. During training, only these LoRA modules, the lightweight spatial adapters, and the Mask2Former decoder are updated. The network is initialized with COCO-panoptic weights to leverage strong geometric priors. Comprehensive architectural details are provided in Appendix~\ref{app:architecture}. Formally, given the query image $I_q \in \mathbb{R}^{3 \times H \times W}$, the network predicts a unified panoptic mask $P_q$, which is decoupled into two distinct sets of masks:
\begin{equation}
  P_q = \mathcal{B}_q \cup \mathcal{R}_q, \quad \mathcal{B}_q = \{M_q^j\}_{j=1}^{N_q}, \quad \mathcal{R}_q = \{U_q^{c}\}_{c \in \mathcal{C}},
\end{equation}
where $M_q^j$ denotes the binary mask of the $j$-th predicted building instance, and $U_q^c$ denotes the mask for `stuff' category $c \in \mathcal{C}$. These outputs structurally correspond to the Panoptic LoD 3D Map $\mathcal{M}$ and enable direct image-space alignment with any rendered pose hypothesis. The entire architecture is optimized end-to-end utilizing the standard Mask2Former panoptic loss, comprising classification, binary mask, and Dice losses, with all training hyperparameters detailed in Appendix~\ref{app:architecture}.

\subsection{Localization Framework}
\label{sec:localization}

The localization stage refines an externally supplied prior pose against the Panoptic LoD 3D Map. We build on the instance-based fine localization framework of LoD-Loc v3~\cite{peng2026lodlocv3} and extend its alignment objective to jointly evaluate building instances and `stuff' regions. Specifically, each candidate pose is scored using building-instance matching and class-wise overlap of road, vegetation, and water masks. Particle-based optimization then searches for a pose that maximizes this joint score.

\noindent \textbf{Building-instance alignment.}
Given a candidate pose $\boldsymbol{\xi}$, the map renderer produces a corresponding hypothesis silhouette $\mathcal{P}_{\boldsymbol{\xi}} = \{\mathcal{B}_{\boldsymbol{\xi}}, \mathcal{R}_{\boldsymbol{\xi}}\}$, comprising rendered building instances $\mathcal{B}_{\boldsymbol{\xi}} = \{M_{\boldsymbol{\xi}}^k\}_{k=1}^{K}$ and `stuff' regions $\mathcal{R}_{\boldsymbol{\xi}} = \{U_{\boldsymbol{\xi}}^{c}\}_{c \in \mathcal{C}}$. 
To effectively handle occlusions, partial observations, and unmapped elements inherently present in aerial views, we match them using an asymmetric one-way objective. For each query building $M_q^j$, we find its best rendered match via the Dice coefficient:
\begin{equation}
d_j^* = \max_k \frac{2|M_q^j \cap M_{\boldsymbol{\xi}}^k|}{|M_q^j|+|M_{\boldsymbol{\xi}}^k|+\epsilon},
\end{equation}
where $\epsilon$ is a small constant preventing division by zero. These optimal instance scores are aggregated using area-aware weights to formulate the overall `thing' score:
\begin{equation}
  S_{\mathrm{b}}(\boldsymbol{\xi}) = \sum_{j=1}^{N_q} \left( \frac{|M_q^j|}{\sum_l |M_q^l| + \epsilon} \right) d_j^*.
\end{equation}
This one-way formulation ensures that every informative query building must be explained by the hypothesis, without penalizing partially visible or over-segmented rendered map instances.

\noindent \textbf{Stuff-region alignment.}
`Stuff' regions lack instance identity but are geometrically indispensable in scenarios such as road networks, vegetation corridors, and water bodies where buildings are sparse or absent. For these regions, we compute a class-wise Intersection over Union for each category $c \in \mathcal{C}$:
\begin{equation}
  S_{c}(\boldsymbol{\xi}) = \frac{|U_q^{c} \cap U_{\boldsymbol{\xi}}^{c}|}{|U_q^{c} \cup U_{\boldsymbol{\xi}}^{c}| + \epsilon}.
\end{equation}

\noindent \textbf{Panoptic alignment score.}
The final localization objective fuses the complementary `thing' and `stuff' terms evaluations into a unified Panoptic alignment score:
\begin{equation}
  S_{\mathrm{P}}(\boldsymbol{\xi}) = \lambda_{\mathrm{b}} S_{\mathrm{b}}(\boldsymbol{\xi}) + \sum_{c \in \mathcal{C}} \lambda_c S_c(\boldsymbol{\xi}),
\end{equation}
where $\lambda_{\mathrm{b}}$ and $\lambda_c$ are corresponding weight parameters. Building structures remain the strongest geometric anchor when present, while the `stuff' terms provide robust complementary support elsewhere. Compared with purely semantic approaches~\cite{zhu2025lod}, the instance term resolves repeated-building ambiguities in dense cities. Compared with purely instance-based methods~\cite{peng2026lodlocv3}, the `stuff' terms provide additional alignment cues in building-less environments when sufficiently distinctive land-cover boundaries are visible.

\noindent \textbf{Prior-guided direct registration.}
\methodname\ directly samples particles around the prior pose $\boldsymbol{\xi}^{p}$. This elegantly circumvents the computationally intensive coarse retrieval stages of earlier pipelines~\cite{zhu2025lod}, thereby significantly reducing overall inference latency and facilitating real-world UAV deployment. The initial swarm is generated via:
\begin{equation}
  \boldsymbol{\xi}_i^{(0)} = \boldsymbol{\xi}^{p} + \Delta\boldsymbol{\xi}_i, \quad i \in \{1, \dots, N_p\},
\end{equation}
where $(0)$ denotes the initial iteration, $\Delta\boldsymbol{\xi}_i$ perturbs translation and yaw, and $N_p$ is the particle count. Each particle is rendered and evaluated by $S_{\mathrm{P}}(\boldsymbol{\xi}_i)$, with its probability weight assigned via softmax:
\begin{equation}
  w_i = \frac{\exp(\tau S_{\mathrm{P}}(\boldsymbol{\xi}_i))}{\sum_{m=1}^{N_p} \exp(\tau S_{\mathrm{P}}(\boldsymbol{\xi}_m))},
\end{equation}
where $\tau$ controls the distribution temperature. High-scoring particles are preserved and re-perturbed across iterations. The pose yielding the highest alignment score across all iterations is selected as the final estimation $\boldsymbol{\xi}^*$. To mitigate local optima, we utilize a dual-swarm strategy~\cite{zhu2025lod,peng2026lodlocv3} that outputs the globally superior candidate from two independent beams.

\begin{figure*}[htbp]
    \centering
    \includegraphics[width=\linewidth]{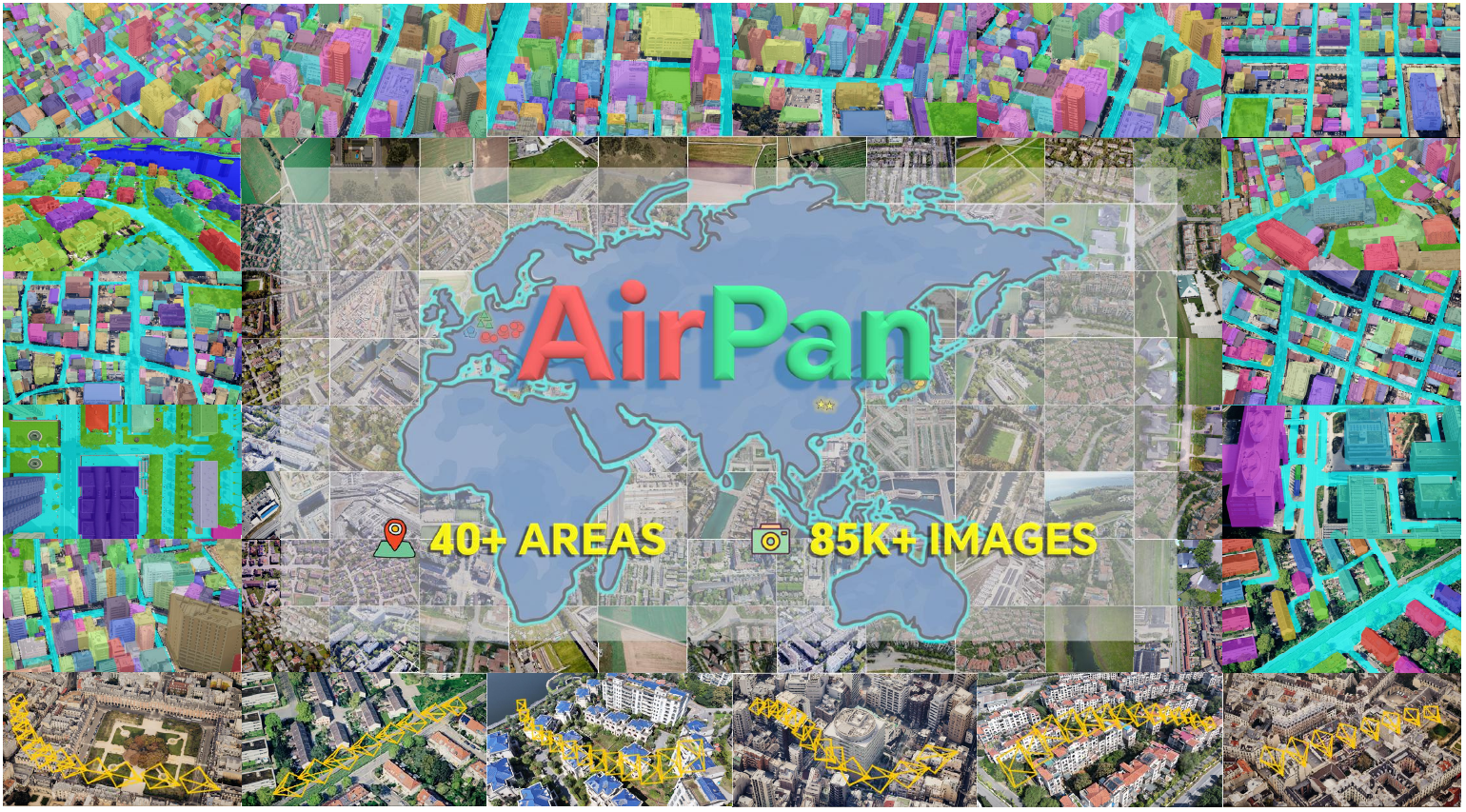}
  \caption{\textbf{Overview of the AirPan dataset.} This figure visualizes representative RGB imagery alongside corresponding ground-truth masks sampled from geographically diverse locations. Encompassing both densely built urban layouts and extremely building-sparse environments, the visualization further summarizes the overall dataset scale in terms of total image volume and spatial coverage.}
  \label{fig:dataset}
\end{figure*}

%% file: section/airpod_dataset.tex
\section{The AirPan Dataset}
\label{sec:dataset}


We introduce \textbf{AirPan}, a massive-scale dataset designed for aerial panoptic segmentation and cross-view localization. As illustrated in Figure~\ref{fig:dataset}, it encompasses diverse urban layouts and building-sparse scenes, summarizing the overall dataset scale in terms of total image volume and spatial coverage. 
To reduce manual annotation costs, we develop a scalable pipeline for data generation and automated panoptic annotation. For the Google Earth-based synthetic imagery, we use Google Earth Studio~\cite{GoogleEarthStudio} to render RGB images along predefined camera trajectories at varying altitudes and viewing angles. For each image, OpenSceneGraph (OSG)~\cite{OSG} renders the corresponding Panoptic LoD 3D Map (Section~\ref{sec:pod-map}) using matching camera poses and projection parameters to generate panoptic masks. Building instances are assigned unique identifier colors, while continuous `stuff' regions use class-specific semantic colors.

\textbf{AirPan} comprises 85,162 images covering 43 distinct geographic regions. To further enhance scene diversity and cross-domain robustness, we curate an auxiliary dataset of 25,419 images from existing open-source benchmarks, including VALID~\cite{valid}, ClaraVid~\cite{claravid}, Aeroscapes~\cite{nigam2018ensemble}, and FloodNet~\cite{rahnemoonfar2021floodnet}, remapping and re-annotating them to fit our panoptic taxonomy.

We first reserve 11 geographic regions for localization testing. We then combine the images from the remaining 32 regions with the auxiliary dataset and split this pool into training and validation sets at a ratio of 80:20. Here, geographic zero-shot means that the localization test regions are excluded from panoptic segmentation training, validation, checkpoint selection, and hyperparameter tuning. Further details of the dataset composition and split are provided in Appendix~\ref{app:dataset}.

\begin{figure*}[htbp]
    \centering
    \includegraphics[width=\linewidth]{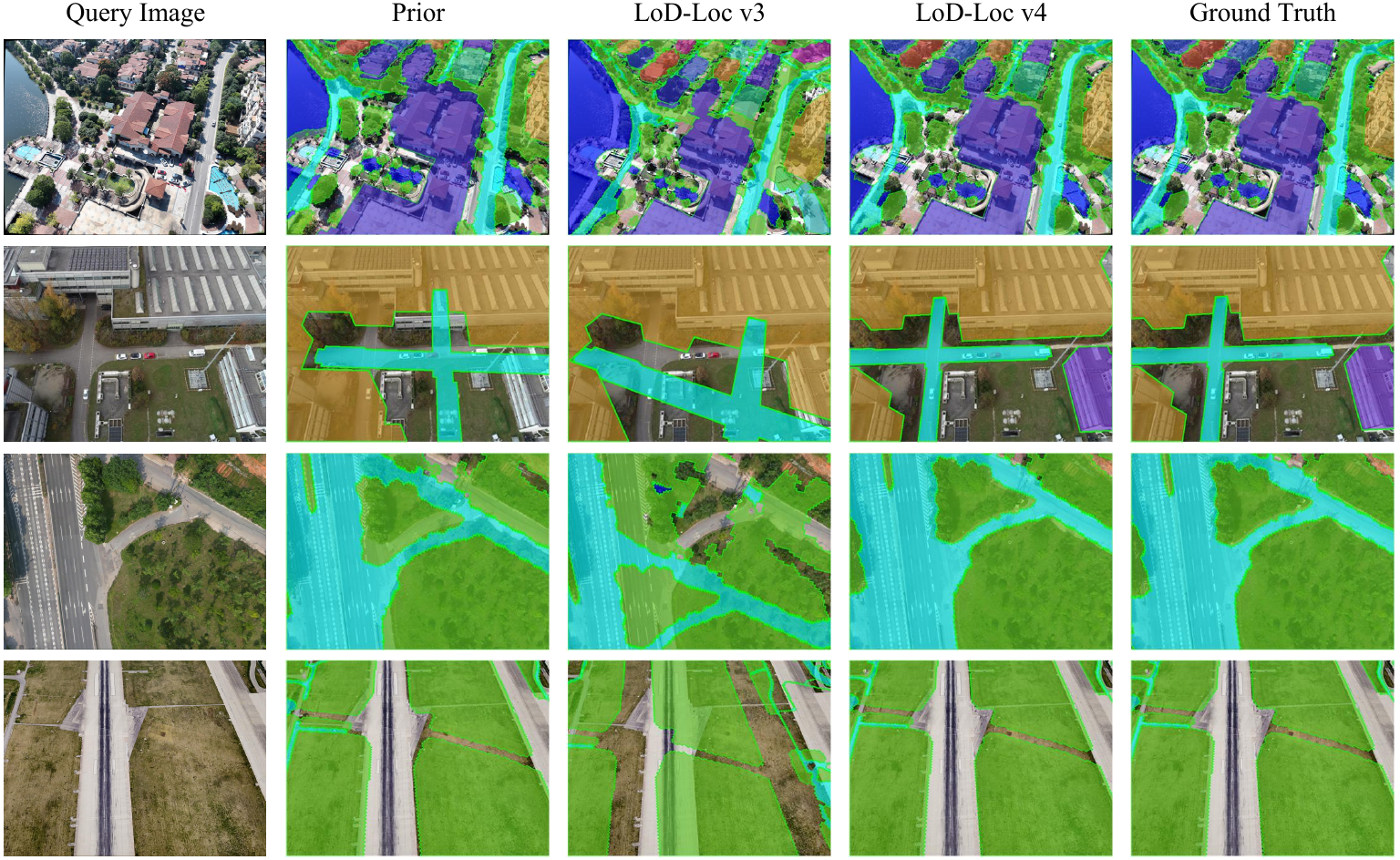}
    \caption{\textbf{Qualitative comparison between LoD-Loc v3 and \methodname.} The columns from left to right show: query image, prior pose, LoD-Loc v3, LoD-Loc v4 (Ours), and the ground truth. All overlays show the same panoptic map rendered at the corresponding poses.}

  \label{fig:visualization}
\end{figure*}

%% file: section/experiments.tex
\section{Experiments}
\label{sec:exp}

In this section, we first describe the experimental setup in Section~\ref{sec:exp_setup}, followed by localization results in Section~\ref{sec:loc_results}, ablation studies in Section~\ref{sec:ab_study}, and qualitative visualizations in Section~\ref{sec:qual_results}.

\subsection{Experimental Setup}
\label{sec:exp_setup}

\noindent \textbf{Benchmarks.}
We evaluate our method on five datasets: UAVD4L-LoDv2~\cite{wu2024uavd4l}, Swiss-EPFLv2~\cite{zhu2025lod}, Tokyo-LoDv3~\cite{peng2026lodlocv3}, China-NB, and Switzerland-NB. UAVD4L-LoDv2 and Swiss-EPFLv2 test the localization performance in real-world scenarios. Tokyo-LoDv3 provides a challenging testbed of dense urban scenes. China-NB and Switzerland-NB serve as the test sets of the \textbf{AirPan} dataset to evaluate localization performance in building-less scenes. Further details are provided in Appendix~\ref{app:experiments}.

\noindent \textbf{Prior pose.}
For UAVD4L-LoDv2, prior poses are obtained from onboard GPS/IMU. For all other test sets, we synthesize priors by adding independent zero-mean Gaussian noise to the ground-truth translation and yaw, with standard deviations of $10\,\mathrm{m}$ per translation axis and $7.5^{\circ}$ for yaw.

\noindent \textbf{Baselines.}
We compare against three LoD-based localization methods, LoD-Loc~\cite{zhu2024lod}, LoD-Loc v2~\cite{zhu2025lod}, and LoD-Loc v3~\cite{peng2026lodlocv3}, and two OSM-based methods, OrienterNet~\cite{sarlin2023orienternet} and VecMapLocNet~\cite{wang2025vecmaplocnet}. All methods are evaluated on the same query images using the same input priors. Further details are provided in Appendix~\ref{app:baselines}.

\noindent \textbf{Evaluation metrics.}
Following standard localization evaluation protocols~\cite{toft2020long}, we report the percentage of queries localized within $(2\text{m}, 2^{\circ})$, $(3\text{m}, 3^{\circ})$, and $(5\text{m}, 5^{\circ})$ thresholds, along with median translation error T.e.\ in meters and median rotation error R.e.\ in degrees.

\textbf{Panoptic segmentation quality.}
For panoptic segmentation, we report the standard Panoptic Quality (PQ), Segmentation Quality (SQ), and Recognition Quality (RQ) on our validation split. Table~\ref{Tab:metrix} reports the class-wise and overall panoptic metrics. The model achieves an overall PQ of 80.47, demonstrating highly balanced performance across all categories. Specifically, the buildings class reaches a PQ of 75.71, while the `stuff' classes achieve PQ scores ranging from 81.30 to 90.86, confirming reliable panoptic predictions for downstream localization.
\input{Table/metrix.tex}

\noindent \textbf{Implementation details.}
The segmentation network is trained for 30 epochs across seven NVIDIA A100 GPUs. Pose estimation inference, employing a dual-swarm particle filter, is executed on a single RTX 4090 GPU. Full details are provided in Appendix~\ref{app:experiments}.

\input{Table/UAVD4L-V2.tex}

\subsection{Localization Results}
\label{sec:loc_results}

\noindent \textbf{Results on UAVD4L-LoDv2.}
As shown in Table~\ref{Tab:uavd4l}, \methodname\ achieves the strongest overall precision for both \textit{in-Traj.}\ and \textit{out-of-Traj.}\ queries. It secures the lowest median translation errors and the highest recall at the ($2\text{m}, 2^{\circ}$) threshold, while remaining highly competitive at looser thresholds, demonstrating strong cross-domain generalization.

\input{Table/Swiss_v2.tex}

\noindent \textbf{Results on Swiss-EPFLv2.}
As reported in Table~\ref{Tab:table_swiss}, \methodname\ establishes a new SOTA for strict-threshold recall and median translation error in both \textit{in-Place.}\ and \textit{out-of-Place.}\ queries. The advantages are particularly pronounced under the \textit{out-of-Place.}\ queries, where our method consistently outperforms prior baselines across all thresholds, highlighting its exceptional spatial transferability.

\input{Table/Tokyo_v3.tex}

\noindent \textbf{Results on Tokyo-LoDv3.}
As presented in Table~\ref{Tab:table_tokyo}, \methodname\ consistently dominates LoD-Loc v3 across all recall metrics and achieves notably lower median translation errors for both \textit{Grid-Traj.}\ and \textit{Sequence-Traj.}\ queries. This demonstrates that incorporating continuous `stuff' semantics effectively resolves severe geometric ambiguities in hyper-dense urban environments.

\input{Table/nobuilding.tex}

\noindent \textbf{Results on China-NB, and Switzerland-NB.}
As shown in Table~\ref{Tab:nobuilding}, all building-based LoD-Loc baselines yield zero recall across all thresholds because these scenes lack building structures to anchor their alignment objectives. OrienterNet also yields zero recall at all thresholds on both datasets, whereas VecMapLocNet obtains zero recall on Switzerland-NB and only 0.74\% recall at the 5m--5$^{\circ}$ threshold on China-NB. In stark contrast, \methodname\ achieves highly reliable localization on both benchmarks. This directly validates our core motivation: `stuff' cues alone can sustain robust localization in building-less scenarios where all building-centric paradigms catastrophically fail.

\subsection{Ablation Studies}
\label{sec:ab_study}

We conduct ablation studies on the Swiss-EPFLv2 and Tokyo-LoDv3 (Japan\_06 and Japan\_07) datasets, focusing on the impact of panoptic composition and segmentation quality.

\noindent \textbf{Panoptic composition.}
As shown in Table~\ref{Tab:ablation_variant}, discarding `thing' cues on the Swiss-EPFLv2 dataset severely degrades recall, confirming building instances as the primary discriminative signal. Conversely, eliminating continuous `stuff' cues also induces notable performance drops, highlighting their crucial complementary role. The full model synergizes both to achieve optimal results across all metrics, validating that `thing'-`stuff' complementarity is indispensable for robust localization.

\input{Table/ablation_variant.tex}

\noindent \textbf{Segmentation quality.}
As reported in Table~\ref{Tab:ablation_checkpoint}, we examine the correlation between panoptic segmentation quality (PQ) and localization accuracy on the Japan\_06 and Japan\_07 queries. Higher PQ generally corresponds to higher recall and lower median translation error, indicating a positive correlation between segmentation quality and pose estimation accuracy.
\input{Table/ablation_checkpoint.tex}

\subsection{Qualitative Results}
\label{sec:qual_results}

Figure~\ref{fig:visualization} presents qualitative visualizations. In urban scenes, \methodname\ achieves significantly tighter alignment than LoD-Loc v3, effectively suppressing geometric ambiguities by integrating surrounding `stuff' context. In building-less scenes, LoD-Loc v3 fails entirely due to the absence of discrete building instances. In stark contrast, \methodname\ successfully drives accurate pose optimization by exploiting continuous road, vegetation, and water layouts. These compellingly illustrate the immense practical value of extending the building-only paradigm to panoptic coverage.

%% file: Table/metrix.tex

\begin{table*}[htbp]
\centering
\caption{\textbf{Panoptic segmentation quality.}}
\label{Tab:metrix}
\begin{tabular}{cccccccc} 
\toprule
Metric & All & Thing & Stuff & Vegetation & Water & Road & Background \\
\cmidrule{1-8}
PQ     & 80.47 & 75.71  & 81.66 & 90.86      & 81.30  & 82.95 & 71.52      \\
SQ     & 88.69 & 87.55  & 88.98 & 93.87      & 93.56  & 85.63 & 82.85      \\
RQ     & 90.67 & 86.48  & 91.72 & 96.80      & 86.90  & 96.87 & 86.32      \\
\toprule
\end{tabular}
\end{table*}

%% file: Table/UAVD4L-V2.tex
\begin{table*}[htbp]
\centering
\setlength{\tabcolsep}{4.2pt} 
\caption{\textbf{Quantitative comparison results of different methods over UAVD4L-LoDv2 dataset.} T.e. and R.e. denote median translation error (m) and median rotation error ($^{\circ}$), respectively.  `-' indicates that the median translation and rotation errors exceed $50\,\mathrm{m}$ and $10^{\circ}$.}
\label{Tab:uavd4l}
\begin{tabular}{ccccccccc} 
\toprule
\multirow{2}{*}{Method} & \multicolumn{4}{c}{\textit{in-Traj.}}                                 & \multicolumn{4}{c}{\textit{out-of-Traj.}}                              \\ 
\cmidrule(r){2-5} \cmidrule(r){6-9}
                        & 2m-2$^{\circ}$          & 3m-3$^{\circ}$          & 5m-5$^{\circ}$          & T.e./R.e.          & 2m-2$^{\circ}$          & 3m-3$^{\circ}$          & 5m-5$^{\circ}$          & T.e./R.e.           \\ 
\cmidrule(r){1-9}
Prior                   & 0              & 0              & 4.30           & 6.48/1.63          & 0              & 0              & 0.36           & 11.10/0.92          \\
OrienterNet~\cite{sarlin2023orienternet}
& 0 & 0 & 0.06 & -
& 0 & 0 & 0 & - \\

VecMapLocNet~\cite{wang2025vecmaplocnet}
& 0 & 0 & 0.06 & -
& 0 & 0 & 0 & - \\
LoD-Loc~\cite{zhu2024lod}                & 49.56          & 71.82          & 89.09          & 3.32/1.48          & 54.20          & 75.05          & 89.51          & 3.33/1.18           \\
LoD-Loc v2~\cite{zhu2025lod}              & 93.70          & 98.40          & 99.50          & 0.72/0.15          & 97.90          & \textbf{99.80}          & \textbf{100.00}         & 0.71/0.14           \\
LoD-Loc v3~\cite{peng2026lodlocv3}              & 97.60          & 98.90          & 99.70          & 0.49/0.13          & 97.40          & 99.00          & 99.40          & 0.60/0.12           \\
\textbf{LoD-Loc v4}     & \textbf{99.60} & \textbf{100.00} & \textbf{100.00} & \textbf{0.42/0.12} & \textbf{99.50} & 99.60 & 99.70 & \textbf{0.40/0.10}  \\
\bottomrule
\end{tabular}
\end{table*}

%% file: Table/Swiss_v2.tex
\begin{table*}[htbp]
\centering
\setlength{\tabcolsep}{5.5pt} 
\caption{\textbf{Quantitative comparison results of different methods over Swiss-EPFLv2 dataset.}}
\label{Tab:table_swiss}
\begin{tabular}{ccccccccc} 
\toprule
\multirow{2}{*}{Method}        & \multicolumn{4}{c}{\textit{in-Place.}}                                                                            & \multicolumn{4}{c}{\textit{out-of-Place.}}                                                                         \\ 
\cmidrule(r){2-5} \cmidrule(r){6-9}
                               & 2m-2$^{\circ}$                     & 3m-3$^{\circ}$                     & 5m-5$^{\circ}$                     & T.e./R.e.                     & 2m-2$^{\circ}$                     & 3m-3$^{\circ}$                     & 5m-5$^{\circ}$                     & T.e./R.e.                      \\ 
\cmidrule(r){1-9}
Prior                   & 0            & 0            & 0.56         & 17.6/3.87 & 0            & 0            & 1.06         & 17.9/3.94  \\
OrienterNet
& 0 & 0 & 0 & -
& 0 & 0 & 0 & - \\

VecMapLocNet
& 0 & 0 & 0 & -
& 0 & 0 & 0 & - \\
LoD-Loc                 & 36.79        & 50.56        & 69.77        & 2.87/1.78 & 14.24        & 31.39        & 59.89        & 8.73/2.78  \\
LoD-Loc v2              & 54.20        & 74.60        & 92.00        & 1.83/0.85 & 31.40        & 58.53        & 86.30        & 2.64/0.73  \\
LoD-Loc v3              & 58.60        & 79.90        & \textbf{95.40}        & 1.61/0.77 & 36.90        & 64.10        & 88.90        & 2.43/0.83  \\
\textbf{LoD-Loc v4}              & \textbf{71.30}        & \textbf{84.10}        & 93.10        & \textbf{1.29/0.61} & \textbf{53.80}        & \textbf{74.10}        & \textbf{91.30}        & \textbf{1.89/0.64}  \\
\toprule
\end{tabular}
\end{table*}


%% file: Table/Tokyo_v3.tex
\begin{table*}[htbp]
\centering
\setlength{\tabcolsep}{5.5pt} 
\caption{\textbf{Quantitative comparison results of different methods over Tokyo-LoDv3 dataset.}}
\label{Tab:table_tokyo}
\begin{tabular}{ccccccccc} 
\toprule
\multirow{2}{*}{Method}                 & \multicolumn{4}{c}{\textit{Grid-Traj.}}                                                                                                               & \multicolumn{4}{c}{\textit{Sequence-Traj.}}                                                                                                            \\ 
\cmidrule(r){2-5} \cmidrule(r){6-9}
                                        & 2m-2$^{\circ}$                              & 3m-3$^{\circ}$                              & 5m-5$^{\circ}$                              & T.e./R.e.                              & 2m-2$^{\circ}$                              & 3m-3$^{\circ}$                              & 5m-5$^{\circ}$                              & T.e./R.e.                               \\ 
\cmidrule(r){1-9}
Prior                   & 0.10           & 1.70           & 8.90           & 8.03/1.78          & 0.30           & 0.80           & 8.80           & 8.19/1.76          \\
OrienterNet
& 0 & 0 & 0 & -
& 0 & 0 & 0 & - \\

VecMapLocNet
& 0.07 & 0.15 & 0.51 & -
& 0 & 0 & 0 & - \\
LoD-Loc                 & 10.52          & 20.95          & 33.80          & 8.03/1.78          & 0              & 0              & 5.06           & 8.19/1.76           \\
LoD-Loc v2              & 2.70           & 8.10           & 22.70          & 7.86/1.48          & 2.30           & 6.80           & 24.00          & 8.75/1.52           \\
LoD-Loc v3              & 39.30          & 68.00          & 89.90          & 2.29/0.27          & 50.30          & 79.90          & 97.30          & 1.98/0.23           \\
\textbf{LoD-Loc v4}     & \textbf{52.85} & \textbf{80.19} & \textbf{95.75} & \textbf{1.92/0.24} & \textbf{58.34} & \textbf{83.46} & \textbf{98.21} & \textbf{1.80/0.23}  \\
\bottomrule
\end{tabular}
\end{table*}

%% file: Table/nobuilding.tex
\begin{table*}[htbp]
\centering
\setlength{\tabcolsep}{5pt} 
\caption{\textbf{Quantitative comparison results of different methods over building-less scenes.}}
\label{Tab:nobuilding}
\begin{tabular}{ccccccccc} 
\toprule
\multirow{2}{*}{Method} & \multicolumn{4}{c}{China-NB}                                             & \multicolumn{4}{c}{Switzerland-NB}                                           \\ 
\cmidrule(r){2-5}\cmidrule(r){6-9}
                        & 2m-2$^\circ$   & 3m-3$^\circ$   & 5m-5$^\circ$ & T.e./R.e.          & 2m-2$^\circ$   & 3m-3$^\circ$   & 5m-5$^\circ$ & T.e./R.e.           \\ 
\cmidrule(r){1-9}
Prior                   & 0              & 0           & 6.32          & 8.29/0.17          & 0           & 0.43           & 3.42          & 11.42/0.17          \\
OrienterNet
& 0 & 0 & 0 & -
& 0 & 0 & 0 & - \\

VecMapLocNet
& 0 & 0.37 & 0.74 & -
& 0 & 0 & 0 & - \\
LoD-Loc                 & 0              & 0              & 0              & -                  & 0              & 0              & 0              & -                   \\
LoD-Loc v2              & 0              & 0              & 0              & -                  & 0              & 0              & 0              & -                   \\
LoD-Loc v3              & 0              & 0              & 0              & -                  & 0              & 0              & 0              & -                   \\
\textbf{LoD-Loc v4}     & \textbf{45.00} & \textbf{57.60} & \textbf{73.60} & \textbf{2.36/0.56} & \textbf{29.90} & \textbf{52.10} & \textbf{78.20} & \textbf{2.91/0.25}  \\
\toprule
\end{tabular}
\end{table*}

%% file: Table/ablation_variant.tex
\begin{table*}[htbp]
\centering
\caption{\textbf{Ablation study on panoptic composition.}}
\label{Tab:ablation_variant}
\begin{tabular}{ccccccccc} 
\toprule
\multirow{2}{*}{Composition} & \multicolumn{4}{c}{\textit{in-Place.}} & \multicolumn{4}{c}{\textit{out-of-Place.}}  \\ 
\cmidrule(r){2-5}\cmidrule(r){6-9}
                         & 2m-2$^{\circ}$      & 3m-3$^{\circ}$      & 5m-5$^{\circ}$      & T.e./R.e.      & 2m-2$^{\circ}$      & 3m-3$^{\circ}$      & 5m-5$^{\circ}$      & T.e./R.e.           \\ 
\cmidrule(r){1-9}
no thing                 & 52.20 & 69.40 & 81.20 & 1.87/0.71      & 37.20 & 63.10 & 84.50 & 2.38/0.66           \\
no stuff                 & 61.00 & 77.10 & 90.20 & 1.51/0.79      & 49.90 & 73.40 & 90.20 & 2.01/0.73           \\
\textbf{Full}                     & \textbf{71.30} & \textbf{84.10} & \textbf{93.10} & \textbf{1.29/0.61}      & \textbf{53.80} & \textbf{74.10} & \textbf{91.30} & \textbf{1.89/0.64}           \\
\toprule
\end{tabular}
\end{table*}

%% file: Table/ablation_checkpoint.tex


\begin{table*}[htbp]
\centering
\caption{\textbf{Ablation study on segmentation quality.}}
\label{Tab:ablation_checkpoint}
\begin{tabular}{cccccccccc} 
\toprule
\multirow{2}{*}{Epoch} & \multirow{2}{*}{PQ} & \multicolumn{4}{c}{Japan\_06}     & \multicolumn{4}{c}{Japan\_07}      \\ 
\cmidrule(r){3-6}\cmidrule(r){7-10}
                       &                     & 2m-2$^{\circ}$   & 3m-3$^{\circ}$   & 5m-5$^{\circ}$   & T.e/R.e   & 2m-2$^{\circ}$   & 3m-3$^{\circ}$   & 5m-5$^{\circ}$   & T.e/R.e    \\ 
\cmidrule(r){1-10}
10                      & 44.38               & 49.50 & 78.00 & 94.90 & 2.02/0.25 & 43.20 & 71.20 & 92.60 & 2.17/0.26  \\
20                     & 53.28              & 53.10 & 80.20 & 95.20 & 1.95/0.25 & 49.00 & 73.30 & 92.30 & 2.04/0.25  \\
\textbf{30}                     & \textbf{63.74}               & \textbf{57.00} & \textbf{85.30} & \textbf{97.60} & \textbf{1.81/0.23} & \textbf{50.50} & \textbf{77.30} & \textbf{94.70} & \textbf{1.99/0.25}  \\
\toprule
\end{tabular}
\end{table*}

%% file: section/conclusion.tex
\section{Conclusion}
\label{sec:conclusion}

This paper presents \methodname, a novel aerial localization paradigm that overcomes the fundamental limitations of traditional building-centric maps. Our solution is twofold: a scalable, highly automated pipeline generates \textbf{AirPan}, a massive aerial panoptic dataset that empowers robust zero-shot perception, while a new registration strategy reformulates 4-DoF pose estimation as a comprehensive alignment over Panoptic LoD 3D Maps. Extensive experiments demonstrate that by jointly leveraging discrete instance-level `thing' and continuous `stuff', \methodname\ not only achieves SOTA precision in dense urban areas but also sustains reliable localization in building-less regions where prior baselines fail, significantly advancing the practicality of global-scale aerial localization.

\noindent \textbf{Limitations and Future Work.} The current formulation localizes each query frame independently without exploiting the temporal consistency across sequential images. In textureless or exceptionally ambiguous regions, this lack of temporal context may occasionally limit pose stability. Incorporating motion models, temporal filtering, or multi-frame aggregation represents a critical direction for future work to seamlessly extend this panoptic alignment paradigm to continuous sequential localization.

\noindent \textbf{Broader Impacts.}
This work can support UAV navigation in GNSS-denied environments. However, improved localization may also facilitate autonomous guidance and aerial surveillance when integrated with sensing, planning, and flight-control systems, raising privacy and misuse concerns. Our method extends existing LoD localization techniques using publicly available geographic data. Its abstract, untextured map representation contains less appearance detail than photorealistic maps, reducing some concerns about disclosing fine visual details. These properties do not eliminate privacy or misuse risks in downstream applications.


%% file: section/appendix.tex
\clearpage
\section{Demo Video}
\label{app:demo_video}

The supplementary material includes a video showcasing the localization performance of \methodname\ across diverse scenes. The video features multiple aerial sequences varying in region, altitude, and viewing angle. We visualize the results by overlaying the projected Panoptic LoD 3D Map onto the query images based on the estimated poses. The sequences cover both building-dense urban areas and building-less scenes comprising only road, vegetation, and water. The visualizations demonstrate that \methodname\ not only achieves SOTA localization precision in dense urban environments but also sustains reliable localization in building-less regions where prior baselines fail.

\section{Architecture of the Segmentation Module}
\label{app:architecture}


This section details the panoptic segmentation architecture designed to parse aerial query images into map-compatible `thing' and `stuff' masks. The network employs a Mask2Former-style framework paired with a BEiT-Adapter-Large backbone, PEFT via LoRA, and a custom panoptic head strictly tailored to our \textbf{AirPan} taxonomy.

\noindent \textbf{Backbone.}
We utilize the BEiT-Adapter-Large backbone~\cite{chen2023vitadapter}, initialized with COCO-Panoptic pretrained weights. The core BEiT encoder comprises 24 transformer blocks with an embedding dimension of 1024 and 16 attention heads. To balance computational efficiency and receptive field size, most layers apply local window attention (window size 14), while every sixth layer employs global window attention (window size 56). The drop-path rate is set to 0.3. Crucially, the core BEiT transformer blocks remain strictly frozen during aerial adaptation to preserve the robust general visual representations acquired during large-scale pretraining.

\noindent \textbf{Adapter module.}
Following ViT-Adapter~\cite{chen2023vitadapter}, lightweight adapter modules are incorporated to inject spatial inductive biases into the frozen ViT representations. Initially, a Spatial Prior Module (SPM) extracts CNN-style multi-scale features at $\{4\times, 8\times, 16\times, 32\times\}$ downsampling rates. These features interact with the BEiT token sequence across four Interaction Blocks (transformer layers [0--5], [6--11], [12--17], and [18--23]). Within each block, an Injector fuses the CNN features into the ViT sequence via multi-scale deformable attention, while an Extractor projects the enriched ViT features back into dense feature maps. The final output head fuses the adapter and ViT features across all four scales, yielding the spatial representations requisite for dense panoptic prediction.

\noindent \textbf{LoRA injection.}
To specialize the pretrained backbone for aerial imagery without the optimization risks and computational overhead of full fine-tuning, we inject LoRA modules~\cite{hu2022lora} into the $\mathrm{QKV}$ projections of all 24 BEiT attention layers. We configure the LoRA rank to $r{=}16$, scaling factor to $\alpha{=}16$, and dropout rate to 0.05. For an input feature $\mathbf{x}$, the adapted projection is formulated as:
\begin{equation}
    \mathbf{y} = \mathrm{QKV}(\mathbf{x}) + \frac{\alpha}{r}\mathbf{B}(\mathbf{A}(\mathbf{x})),
\end{equation}
where $\mathbf{A} \in \mathbb{R}^{1024 \times 16}$ and $\mathbf{B} \in \mathbb{R}^{16 \times 3072}$ denote the low-rank trainable matrices. During training, only the LoRA parameters, adapter modules, and Mask2Former heads are optimized, while the core BEiT blocks are frozen ($\text{lr\_mult}{=}0$). This PEFT strategy drastically reduces the trainable parameter count, mitigating the risk of overfitting on the aerial domain while maintaining the network's general representation capacity.

\noindent \textbf{Panoptic head.}
We employ a Mask2Former panoptic head~\cite{cheng2022masked} configured with 100 learnable queries. The pixel decoder features a 6-layer multi-scale deformable-attention encoder with an embedding dimension of 256, 8 attention heads, 3 feature levels, and 4 sampling points. The transformer decoder comprises 9 layers with an embedding dimension of 256, 8 attention heads, and a feedforward dimension of 2048. Each query independently predicts a class label and a binary mask. These predictions are subsequently fused into panoptic outputs spanning five categories perfectly aligned with our map taxonomy: one `thing' class (buildings), three `stuff' classes (road, vegetation, water), and one background class. The background class is supervised during training but deliberately discarded by the downstream localization objective; additionally, unlabeled pixels (index 255) are excluded from all loss computations. The training objectives follow the standard Mask2Former protocol, employing a classification loss weight of 2.0, a binary mask loss weight of 5.0, a Dice loss weight of 5.0, and identical weights for the Hungarian matching costs.

\section{Details of the AirPan Dataset}
\label{app:dataset}

\subsection{Geographic Data Sources}
\label{app:data_sources}

The Panoptic LoD 3D Map is assembled from three complementary public data modalities: OSM vectors for semantics, DEM tiles for terrain elevation, and LoD building models for 3D geometry. This subsection details the acquisition and standardization of each modality prior to map fusion.

\noindent \textbf{OpenStreetMap (OSM) vector data.}
We utilize the Overpass API to retrieve vector layers from OSM~\cite{OSM} within each target bounding box, encompassing road networks, vegetation polygons, water bodies, and coastlines. Our acquisition pipeline implements parallel requests and rate-limiting protocols, ensuring efficient batch processing across large-scale regions. The retrieved geometries are filtered by spatial tags and categorically mapped into three `stuff' classes: road, vegetation, and water. All remaining unassigned polygons are classified as background.

\noindent \textbf{Digital Elevation Models (DEM).}
Terrain elevation is sourced from region-specific official DEM products to preserve the highest available topographical resolution. Specifically, we utilize swissALTI3D~\cite{swissalti3d} at 2\,m resolution for Switzerland, AHN3/AHN4~\cite{ahn} at 0.5\,m for the Netherlands, the GSI DEM~\cite{gsi_dem} at 5\,m or 10\,m for Japan, Copernicus DEM~\cite{copernicus_dem} at 30\,m for Italy, and RGE ALTI~\cite{rge_alti} at 5\,m for France. All DEM tiles are acquired in GeoTIFF format, cropped to the target bounding box, and spatially aligned with the rasterized semantic map prior to terrain reconstruction.

\noindent \textbf{LoD building models.}
LoD building models are collected from national and municipal 3D city model repositories. We employ PLATEAU~\cite{PLATEAU} for Japan, swissBUILDINGS3D~\cite{swisstopo} for Switzerland, and 3D-BAG~\cite{3D_BAGS} for the Netherlands; for other European regions, we rely on open 3D model resources following the protocols of prior work~\cite{zamir2016generic}. All building models are converted into OBJ meshes and transformed into the EPSG:4978 geocentric reference frame, ensuring strict spatial consistency with the DEM surface and the downstream rendering engine.

\subsection{Map Construction Details}
\label{app:map_construction}

The proposed automated pipeline converts raw geographic data into a renderable Panoptic LoD 3D Map through four sequential stages. 

\noindent \textbf{Stuff-region rasterization.}
We begin by rasterizing OSM land-use polygons into a semantic map at zoom level 18, which provides a practical balance between spatial detail and processing cost. Each pixel is then deterministically assigned a predefined semantic color corresponding to road, vegetation, water, building footprint, or background. 

\noindent \textbf{2.5D terrain construction.}
Subsequently, the rasterized map is elevated into a 3D structure utilizing the aligned DEM. Raster coordinates are converted from Web Mercator back to WGS84 and further transformed into a local projected coordinate system via \texttt{pyproj}. Elevation values are sampled from the DEM to construct a triangulated 2.5D terrain surface with metric heights.

\noindent \textbf{Building instance injection.}
Following the instanced LoD pipeline~\cite{peng2026lodlocv3}, we instantiate the LoD building models via topological structure analysis before anchoring them onto the terrain. The untextured LoD mesh is partitioned into connected components, each treated as a discrete building instance. Each instance is assigned a unique 24-bit integer identifier, deterministically mapped to a unique RGB color via the bijective encoding:
\begin{equation}
    \mathbf{c}_i = \Phi(id_i) =
    \begin{pmatrix}
        \lfloor id_i \cdot 256^{-2} \rfloor \\
        \lfloor id_i \cdot 256^{-1} \rfloor \pmod{256} \\
        id_i \pmod{256}
    \end{pmatrix},
\end{equation}
supporting up to 16.7 million unique instances. The resulting colored building instances are injected into the terrain model and merged with the `stuff' surface into a unified OSGB scene.

\noindent \textbf{Coordinate alignment.}
We express the Panoptic LoD 3D Map and the camera poses from Google Earth Studio in the common EPSG:4978 geocentric frame. OSG then renders the map using matching camera intrinsics and extrinsics to generate the corresponding panoptic masks.

\subsection{Annotation Pipeline}
\label{app:rendering_pipeline}


\noindent \textbf{RGB image acquisition.}
We use Google Earth Studio~\cite{GoogleEarthStudio} to generate synthetic aerial RGB sequences along predefined camera trajectories with varying altitudes and viewing angles. For each frame, we retain the camera pose, field of view, and image resolution for subsequent label rendering.

\noindent \textbf{Panoptic annotation.}
We generate panoptic masks by rendering the corresponding Panoptic LoD 3D Map with OpenSceneGraph~\cite{OSG}. Camera poses and map geometry are expressed in a common geographic coordinate frame, and the rendering uses camera projection parameters and image resolution matching those of the RGB frames.

\subsection{Dataset Composition Details}
\label{app:dataset_composition}


\noindent \textbf{AirPan dataset.}
To the best of our knowledge, \textbf{AirPan} constitutes the largest aerial panoptic segmentation dataset currently available, containing 85,162 images. We reuse the camera poses and pose generation procedure from LoD-Loc v3~\cite{peng2026lodlocv3} and re-render the corresponding RGB images using Google Earth Studio. The dataset covers 43 UAV flight areas across Japan, Switzerland, China, France, Italy, and the Netherlands. The resulting data spans diverse land-use types, including commercial, industrial, residential, educational, and suburban regions. Table~\ref{Tab:dataset_region_details} provides a comprehensive breakdown of each geographical area within the dataset, detailing its name, image count, flight altitude range in meters, total geographical area in square kilometers, image resolution in pixels, and captured viewing angles in degrees. In addition, we render two building-less test sets, China-NB and Switzerland-NB, specifically to evaluate localization performance in building-less scenes.
\input{Table_Sup/dataset_region_details}

To further enhance scene diversity and cross-domain robustness, we integrate four additional open datasets to construct an auxiliary set. After sample filtering, semantic label remapping, connected-component analysis for building instances, and manual correction, this auxiliary set contributes 25,419 images. Specifically, VALID~\cite{valid} and ClaraVid~\cite{claravid} provide 6,690 and 16,316 synthetic aerial panoptic segmentation images, respectively. Aeroscapes~\cite{nigam2018ensemble} and FloodNet~\cite{rahnemoonfar2021floodnet} provide 896 and 1,517 real-world aerial semantic segmentation images, respectively.

\noindent \textbf{Data augmentation and splitting.}
We first hold out 11 geographic regions for localization testing: China\_01--02, Switzerland\_17--18, Japan\_06--10, China-NB, and Switzerland-NB. Japan\_09* is an adverse-weather variant of the held-out Japan\_09 region and is also reserved for testing.

The remaining 32 regions comprise Netherlands\_01--06, Switzerland\_01--16, Japan\_02--05, Japan\_11, Italy\_01--02, and France\_01--03. Only the images from these regions, together with the 25,419 auxiliary images, are partitioned into training and validation sets at a ratio of 80:20.
During training, we apply random horizontal flipping, multi-scale resizing to four resolutions ($720 \times 480$, $896 \times 608$, $1024 \times 768$, and $1200 \times 896$), and random $768 \times 768$ cropping. Images larger than the crop size are further divided into overlapping patches, yielding approximately 110,906 training samples after patch extraction.

\section{Details of Baselines}
\label{app:baselines}

We evaluate our method against three LoD-based localization baselines and two OSM-based baselines. All methods use the same query images and prior poses. For OrienterNet and VecMapLocNet, we use the released checkpoints. Both methods estimate planar location and yaw; height, roll, and pitch are retained from the prior pose.

\noindent \textbf{LoD-Loc~\cite{zhu2024lod}.} The implementation follows its default settings. Specifically, we utilize uniform sampling counts of 10, 10, 30, and 8 along the $x$, $y$, $z$, and yaw axes, respectively, alongside a weighting parameter $\lambda = 0.8$. To adapt the method to our experimental setup, the primary modification involves deploying the Panoptic LoD 3D Map with a 3D sampling interval of 1 meter.

\noindent \textbf{LoD-Loc v2~\cite{zhu2025lod}.} The implementation follows its default settings. The input image resolution for the segmentation module is resized to $1024 \times 1024$. During the coarse pose selection stage, the mask resolutions for both query and rendered candidates are configured to $640 \times 360$, $602 \times 448$, or $720 \times 480$ (depending on the specific dataset), utilizing a sampling interval of 10 meters. The subsequent fine pose estimation stage uses the same mask resolutions and runs for 40 iterations with 2 beams. For particle generation, angular perturbations are constrained to $2^{\circ}$, while translational perturbations are sampled from a zero-mean Gaussian distribution $\mathcal{N}(0, \sigma^2)$ with $\sigma = 1.5$. A decay factor of $\gamma = 0.3$ is applied to the perturbations, generating $n = 52$ candidate poses per iteration.

\noindent \textbf{LoD-Loc v3~\cite{peng2026lodlocv3}.} The implementation follows its default settings, with the input image resolution for the segmentation module set to $512 \times 512$. All other hyperparameter configurations for both the coarse pose selection and fine pose estimation stages are identical to those described for LoD-Loc v2.

\noindent \textbf{OrienterNet~\cite{sarlin2023orienternet}.}
We use \texttt{orienternet\_mgl.ckpt} with 256 yaw bins over $360^{\circ}$ and restrict the search to $\pm15^{\circ}$ around the prior yaw. Each query uses a prior-centered $128\,\mathrm{m}\times128\,\mathrm{m}$ OSM patch at $2$ pixels per meter. Images are rectified to a pure-nadir view using the prior orientation, resized according to the camera intrinsics and checkpoint configuration, and padded to multiples of 32, with intrinsics and validity masks updated accordingly.

\noindent \textbf{VecMapLocNet~\cite{wang2025vecmaplocnet}.}
We use \texttt{last-step-checkpointing-v1.ckpt} with 64 yaw bins searched over the full $360^{\circ}$ range. Each query uses a prior-centered $512\,\mathrm{m}\times512\,\mathrm{m}$ OSM vector-map patch at $1$ pixel per meter. Images are resized to a shorter side of 256 pixels while preserving the aspect ratio.

\begin{figure*}[ht]
  \centering
  \includegraphics[width=1.0\linewidth]{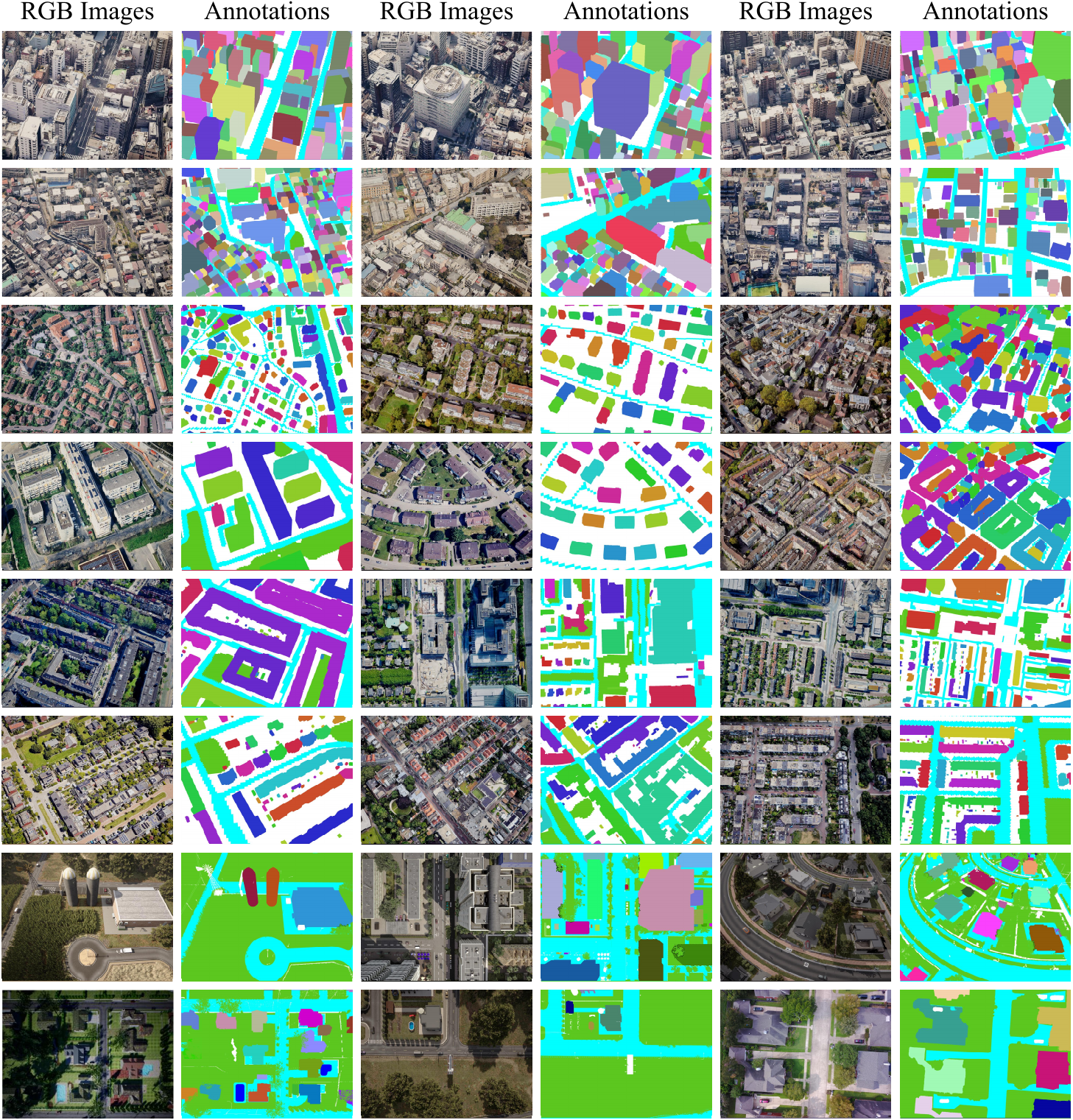}
  \caption{\textbf{Visualization of training samples.} This figure displays pairs of synthetic RGB images and their corresponding panoptic annotations. The top four rows showcase samples of the Japan, Switzerland, and Netherlands regions from the \textbf{AirPan} dataset, while the bottom two rows feature samples from the integrated auxiliary dataset.}
  \label{fig:training_data_visualization}
\end{figure*}

\section{Details of Experiments}
\label{app:experiments}

\begin{figure*}[htbp]
  \centering
  \includegraphics[width=1.0\linewidth]{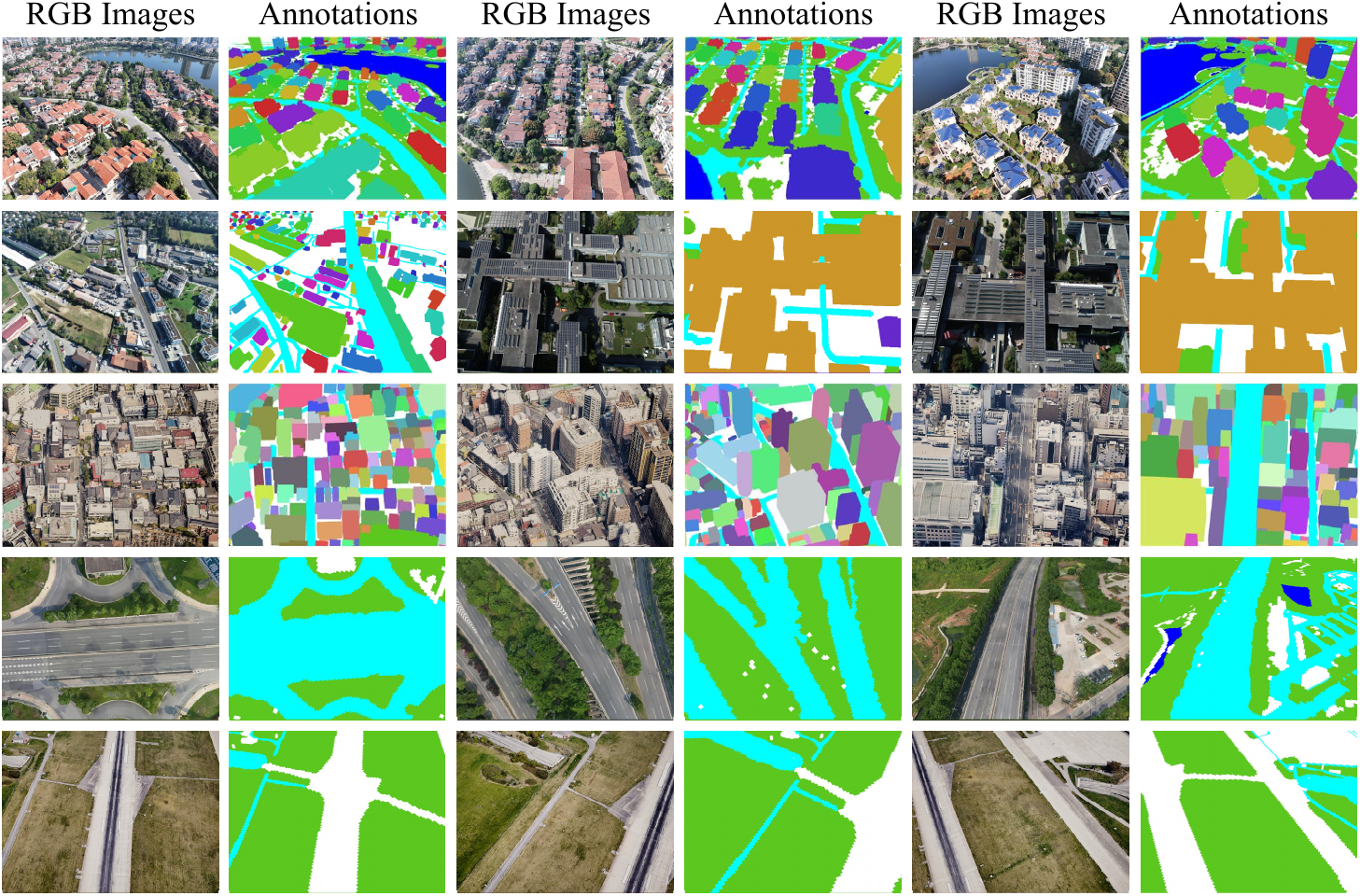}
  \caption{\textbf{Visualization of samples from the test datasets.} Rows show samples from UAVD4L-LoDv2, Swiss-EPFLv2, Tokyo-LoDv3, China-NB, and Switzerland-NB.}
  \label{fig:test_data_visualization}
\end{figure*}

\subsection{Implementation Details}
\label{app:impl_details}

\noindent \textbf{Training.} Our panoptic segmentation module is trained on the large-scale training split containing approximately $110,906$ images (including sliced patches). To enhance scale invariance and scene diversity, we adopt a multi-scale training strategy where input images are dynamically resized to four varying scales ($720 \times 480$, $896 \times 608$, $1024 \times 768$, and $1200 \times 896$), coupled with random crops of $768 \times 768$. To optimize the training process, we use the AdamW optimizer ($\beta_{1}{=}0.9$, $\beta_{2}{=}0.999$) with a base learning rate of $4 \times 10^{-4}$ and a weight decay of $0.05$. We employ a cosine annealing schedule featuring a linear warmup phase of $1000$ iterations (warmup ratio of $0.001$) and a minimum learning rate of $10^{-7}$. The network is trained for $30$ epochs with a batch size of $1$ per GPU. To further stabilize the training dynamics, we apply gradient clipping with a maximum norm of $0.01$ and utilize an Exponential Moving Average (EMA) with a momentum of $0.0001$. The network architecture utilizes a \texttt{BEiT-Adapter-Large} backbone coupled with a Mask2Former head. For parameter-efficient adaptation, the core backbone remains frozen while LoRA modules ($r{=}16$, $\alpha{=}16$) are injected into the attention layers, alongside tunable adapter structures. All training procedures were performed on seven NVIDIA A100 GPUs.

\noindent \textbf{Inference.} Inference is evaluated on test sets differing in region and viewpoint from the training set. During the perception phase, the input size is dynamically adapted to each dataset's native resolution (e.g., $720 \times 480$ or $640 \times 360$). We apply a minimum area of $10$ pixels to filter building instances. In the pose optimization stage, consistent with LoD-Loc v2~\cite{zhu2025lod}, we set the number of iterations to $40$, using $2$ beams and an angular perturbation of $2^{\circ}$. Translational perturbations are sampled from a zero-mean Gaussian distribution $\mathcal{N}(0, \sigma^2)$ with $\sigma = 1.5$. The decay factor is set to $\gamma = 0.3$, and $n = 52$ candidates are generated per iteration. Crucially, to accommodate the proposed panoptic framework, the multi-class silhouette alignment weights are set to $\lambda_b{=}0.6$ for buildings, $\lambda_r{=}0.2$ for road, $\lambda_v{=}0.1$ for vegetation, and $\lambda_w{=}0.1$ for water. All perception inference and localization processes were executed on a single NVIDIA RTX 4090 GPU.

\subsection{Visualization of Training Data}
\label{app:train_vis}


Our training set comprises RGB images and their corresponding panoptic masks, derived from the designated training splits of the \textbf{AirPan} dataset and the integrated auxiliary dataset. It encompasses diverse geographic regions and land-use types to yield a total of 110,906 sliced samples. Figure~\ref{fig:training_data_visualization} presents a visualization of selected samples from the training dataset. In the panoptic annotations, building instances are rendered with unique random colors to denote distinct entities, whereas continuous `stuff' classes, such as road, vegetation, and water bodies, are depicted using fixed semantic colors.

\subsection{Visualization of Test Data}
\label{app:test_vis}

As illustrated in Figure~\ref{fig:test_data_visualization}, the test data consists of five benchmarks:
\begin{itemize}[leftmargin=1.2em]
    \item \textbf{UAVD4L-LoDv2}~\cite{wu2024uavd4l}: A real-world dataset covering a 2.5~km$^2$ area. It comprises two sub-regions (China\_01 and China\_02) and features \textit{in-Traj.}\ and \textit{out-of-Traj.}\ queries.
    \item \textbf{Swiss-EPFLv2}~\cite{zhu2025lod}: A real-world dataset covering an 8.2~km$^2$ area. It comprises two sub-regions (Switzerland\_17 and Switzerland\_18), featuring \textit{in-Place.}\ and \textit{out-of-Place.}\ queries.
    \item \textbf{Tokyo-LoDv3}~\cite{peng2026lodlocv3}: A synthetic dataset covering a 4.06~km$^2$ area of dense urban Tokyo. It contains five distinct sub-regions (Japan\_06 through Japan\_10), categorized into \textit{Grid-Traj.}\ and \textit{Sequence-Traj.}\ queries.
    \item \textbf{China-NB}: A synthetic building-less benchmark covering a 1.2~km$^2$ suburban area of Changsha, featuring road, vegetation, and water bodies.
    \item \textbf{Switzerland-NB}: A synthetic building-less benchmark covering a 2.0~km$^2$ suburban area of Geneva, featuring road and vegetation.
\end{itemize}

We evaluate geographic zero-shot generalization: all localization test regions are excluded from panoptic segmentation training, validation, checkpoint selection, and hyperparameter tuning.

\input{Table/adverse_weather.tex}

\subsection{Supplementary Experiments}
\label{app:adverse_weather}

To further evaluate the robustness of our method in more complex environments, we conduct localization experiments on Japan\_09* with simulated adverse weather effects, including haze, overcast skies, and seasonal appearance changes.

Table~\ref{Tab:adverse_weather_results} reports the results. \methodname\ consistently improves over LoD-Loc v3 across all recall thresholds and reduces median translation error. This demonstrates that `stuff' region boundaries remain structurally stable even when building textures degrade under adverse conditions.


%% file: Table_Sup/dataset_region_details.tex
\begin{table*}[ht]
\centering
\setlength{\tabcolsep}{3pt} 
\caption{\textbf{Detailed characteristics of the regional datasets composing the \textbf{AirPan} dataset.}}
\label{Tab:dataset_region_details}
\begin{tabular}{ccccccc} 
\toprule
Name            & Area      & Images & Height[$m$] & Coverage[$km^2$] & Size[$pixel$] & View Angle[°]  \\ 
\midrule
Netherlands\_01 & Amsterdam & 392   & 500         & 18.3879            & 1600×1200     & {[}0, 30]      \\
Netherlands\_02 & Amsterdam & 1242   & 200         & 17.3980            & 1600×1200     & {[}30, 50]     \\
Netherlands\_03 & Hague     & 2048   & 300         & 16.9807            & 1600×1200     & {[}0, 30]      \\
Netherlands\_04 & Delft     & 953   & 200         & 20.1190            & 1600×1200     & {[}0, 30]      \\
Netherlands\_05 & Amsterdam & 558   & 500         & 16.9844            & 1600×1200     & {[}30, 50]     \\
Netherlands\_06 & Tilburg   & 2874   & 200         & 19.7211            & 1600×1200     & {[}0, 30]      \\
Switzerland\_01 & Basel     & 4771   & 300, 400    & 1.7458             & 1920×1080     & 0, 45          \\
Switzerland\_02 & Ecublens  & 13563  & 400, 500    & 8.6432             & 1920×1080     & 0, 45          \\
Switzerland\_03 & Zurich    & 1783   & 300         & 13.0968            & 1600×1200     & {[}50, 70]     \\
Switzerland\_04 & Zurich    & 3209   & 200         & 17.9719            & 1600×1200     & {[}0, 30]      \\
Switzerland\_05 & Lausanne  & 2269   & 200         & 8.8370             & 1600×1200     & {[}30, 50]    \\
Switzerland\_06 & Geneva    & 1916   & 200         & 17.9286            & 1600×1200     & {[}0, 30]     \\
Switzerland\_07 & Bern      & 3209   & 300         & 15.9526            & 1600×1200     & {[}30, 50]    \\
Switzerland\_08 & Bern      & 1043   & 200         & 17.5524            & 1600×1200     & {[}50, 70]     \\
Switzerland\_09 & Lucerne   & 2665   & 200         & 15.4045            & 1600×1200     & {[}0, 30]      \\
Switzerland\_10 & Zurich    & 3179   & 200         & 17.2618            & 1600×1200     & {[}30, 50]     \\
Switzerland\_11 & Geneva    & 3202   & 200         & 2.7146             & 1600×1200     & {[}30, 50]     \\
Switzerland\_12 & Geneva    & 3210   & 300         & 0.7120             & 1600×1200     & {[}0, 30]      \\
Switzerland\_13 & Geneva    & 3209   & 200         & 13.5406            & 1600×1200     & {[}0, 30]      \\
Switzerland\_14 & Bill      & 3078   & 200         & 17.2042            & 1600×1200     & {[}0, 30]      \\
Switzerland\_15 & Lausanne  & 4300   & 500         & 10.1067            & 1600×1200     & {[}0, 30]      \\
Switzerland\_16 & Lausanne  & 742    & 600, 700    & 0.8579             & 1920×1080     & 0, 45          \\
Switzerland\_17 & Lausanne  & 712    & 400, 500    & 3.2468             & 720×480       & {[}0, 30]      \\
Switzerland\_18 & Lausanne  & 379    & 400, 500    & 5.1028             & 720×480       & {[}0, 30]      \\
Switzerland-NB  & Geneva    & 234    & 200         & 1.2051             & 1600×1200     & {[}0, 30]     \\
Japan\_02       & Tokyo     & 1440   & 350, 450    & 0.4028             & 1920×1080     & 0, 45          \\
Japan\_03       & Tokyo     & 3260   & 350, 450    & 0.9747             & 1920×1080     & 0, 45          \\
Japan\_04       & Tokyo     & 682    & 350, 450    & 0.7171             & 1920×1080     & 0, 45          \\
Japan\_05       & Tokyo     & 1000   & 350, 450    & 0.3302             & 1920×1080     & 0, 45          \\
Japan\_06       & Tokyo     & 495    & 350, 450    & 0.6754             & 1920×1080     & 0, 45          \\
Japan\_07       & Tokyo     & 875    & 350, 450    & 0.3283             & 1920×1080     & 0, 45          \\
Japan\_08       & Tokyo     & 750    & 450, 500    & 1.3225             & 1920×1080     & 0, 45          \\
Japan\_09       & Tokyo     & 381    & 450, 500    & 0.8281             & 1920×1080     & 0, 45          \\
Japan\_09*      & Tokyo     & 381    & 450, 500    & 0.8281             & 1920×1080     & 0, 45          \\
Japan\_10       & Tokyo     & 450    & 450, 500    & 0.9025             & 1920×1080     & 0, 45          \\
Japan\_11       & Tokyo     & 450    & 450, 500    & 0.8464             & 1920×1080     & 0, 45          \\
China\_01       & Changsha  & 1604   & 120         & 1.6384             & 4056×3040     & 0, 45          \\
China\_02       & Changsha  & 2192   & {[}90, 150] & 1.4884             & 5280×3956     & {[}30, 60]     \\
China-NB        & Changsha  & 269    & 120         & 1.1768             & 1204×896      & 0, 45          \\
Italy\_01       & Florence  & 155    & 400, 500    & 0.8778             & 1920×1080     & 0, 45          \\
Italy\_02       & Florence  & 2631   & 400, 500    & 0.1571             & 1920×1080     & 0, 45          \\
France\_01      & Florence  & 421    & 400, 500    & 0.4865             & 1920×1080     & 0, 45          \\
France\_02      & Paris     & 222    & 400, 500    & 0.2751             & 1920×1080     & 0, 45          \\
France\_03      & Paris     & 2764   & 400, 500    & 0.7890             & 1920×1080     & 0, 45          \\
\toprule
\end{tabular}
\end{table*}

%% file: Table/adverse_weather.tex
\begin{table}[htbp]
\centering
\caption{\textbf{Quantitative comparison of different methods under adverse-weather condition.}}
\label{Tab:adverse_weather_results}
\begin{tabular}{ccccc} 
\toprule
\multirow{2}{*}{Method} & \multicolumn{4}{c}{Japan\_09*}                                         \\ 
\cmidrule(r){2-5}
                        & 2m-2$^{\circ}$         & 3m-3$^{\circ}$             & 5m-5$^{\circ}$             & T.e./R.e.             \\
\cmidrule(r){1-5}
Prior                   & 0.52           & 0.79           & 7.35           & 8.14/1.80           \\
LoD-Loc v2              & 1.80           & 3.10           & 8.70           & 14.69/0.99          \\
LoD-Loc v3              & 24.70          & 55.10          & 93.70          & 2.86/0.23           \\
\textbf{LoD-Loc v4}     & \textbf{52.00} & \textbf{81.40} & \textbf{96.90} & \textbf{1.97/0.25}  \\
\toprule
\end{tabular}
\end{table}